\documentclass{article}

\usepackage{mdframed}
\usepackage{listings}
\usepackage{xcolor} 
\usepackage{lipsum} 

    \usepackage[preprint]{neurips_2024}

\usepackage{hyperref}

\usepackage[utf8]{inputenc} 
\usepackage[T1]{fontenc}    
\usepackage{hyperref}       
\usepackage{url}            
\usepackage{booktabs}       
\usepackage{amsfonts}       
\usepackage{nicefrac}       
\usepackage{microtype}      
\usepackage{xcolor}         
 \usepackage{amsmath}
\usepackage{graphicx}  

\title{LLMs can Schedule}

\author{%
  Henrik Abgaryan\thanks{LAMSADE, Université Paris Dauphine - PSL, CNRS, Paris, France. } \\
  Department of Computer Science\\
  Université Paris Dauphine - PSL\\
  Paris, France \\
  \texttt{henrik.abgaryan@dauphine.eu} \\
  \And
  Ararat Harutyunyan \\
  Department of Computer Science\\
  Université Paris Dauphine - PSL\\
  Paris, France \\
  \texttt{ararat.harutyunyan@dauphine.eu} \\
  \And
  Tristan Cazenave \\
  Department of Computer Science\\
  Université Paris Dauphine - PSL\\
  Paris, France \\
  \texttt{tristan.cazenave@dauphine.eu} \\
}

\begin{document}

\maketitle

\begin{abstract}
Abstract: The job shop scheduling problem (JSSP) remains a significant hurdle in optimizing production processes. This challenge involves efficiently allocating jobs to a limited number of machines while minimizing factors like total processing time or job delays. While recent advancements in artificial intelligence have yielded promising solutions, such as reinforcement learning and graph neural networks, this paper explores the potential of Large Language Models (LLMs) for JSSP. We introduce the very first supervised 120k dataset specifically designed to train LLMs for JSSP. Surprisingly, our findings demonstrate that LLM-based scheduling can achieve performance comparable to other neural approaches. Furthermore, we propose a sampling method that enhances the effectiveness of LLMs in tackling JSSP.
\end{abstract}


\section{Introduction}

The job shop scheduling problem (JSSP) remains a well-studied and computationally challenging problem in the field of production scheduling and optimization. It entails the efficient allocation of a set of $N$ jobs, each with heterogeneous processing times, to a limited number of $M$ machines. The primary objective is to optimize a performance metric, such as minimizing the total completion time (makespan, denoted by $C_{max}$) or reducing the flow time (average completion time) of individual jobs. JSSP finds application in diverse manufacturing and service environments, impacting factors like production throughput, resource utilization, and ultimately, customer service levels.
Traditional approaches to JSSP have primarily relied on mathematical programming techniques and heuristic algorithms \cite{performance_of_HeuristicDispatchingRules}. However, these methods often exhibit limitations in scalability and effectiveness, particularly for large-scale problems, or those with complex job-machine precedence relationships. This has motivated the exploration of alternative approaches, particularly with the recent advancements in artificial intelligence (AI). Techniques like reinforcement learning and graph neural networks have shown promise in addressing JSSP, offering data-driven solutions to this problem\cite{L2D}\cite{SelfJSP}.

This paper explores the potential of Large Language Models (LLMs) in tackling the complexities of the job shop scheduling problem (JSSP). To the best of our knowledge, we are the first to utilize large language models (LLMs) for end-to-end scheduling in JSSP problems. On small-scale JSSP problems we show that minimal fine-tuning through LoRA method \cite{hu2022lora} enables LLM to schedule, by finding high-quality solutions, sometimes matching or surpassing traditional neural approaches.
LLMs represent a class of AI models trained on massive datasets of text data. This extensive training equips them with the ability to comprehend and generate human-like language. We posit that LLMs, with their inherent ability to process and reason over complex information, can be effectively harnessed to address JSSP. To this end, we introduce a novel supervised dataset \footnote{\url{https://github.com/starjob42/datasetjsp}} designed to train LLMs specifically for the task of job shop scheduling. This dataset is different from the usual ones used for JSSP problems. Instead of traditional matrix representation format, this dataset includes natural language description of the JSSP problem and solution, specifically made for training LLMs. 

This paper presents a comparative analysis that demonstrates the efficacy of LLM-based scheduling. Our findings indicate that LLMs have the potential to effectively schedule tasks, demonstrating performance on par with some of the current neural network methods for JSSP. Furthermore, we add sampling method on top of the LLM that serves to enhance the effectiveness of LLMs in tackling this problem.
The contributions of this work to the field of JSSP are multifaceted:
\begin{itemize}

    \item We introduce the very first supervised 120k dataset specifically designed for LLM training for JSSP. This dataset addresses the unique requirements of the problem domain, facilitating effective LLM training.
    \item We explore the potential application of LLMs for job shop scheduling, offering a novel approach.
    \item We present a comparative analysis demonstrating the effectiveness of end-to-end LLM-based scheduling compared to existing neural network approaches. 
    \item This analysis sheds light on the capabilities of LLMs in this domain.
    \item We use sampling method that improves LLM performance in JSSP.
\end{itemize}

In summary, this paper introduces a new supervised dataset for LLM training for JSSP, offering valuable insights and a robust framework for further exploration in this area.

\section{Related Work}

The job shop scheduling problem (JSSP) with more than two machines is proven to be NP-hard \cite{garey1976complexity}. As a result, finding exact solutions for JSSP is generally infeasible, leading to the widespread use of heuristic and approximate methods for practical efficiency \cite{jssp_np}.
Traditional approaches to solving JSSP have primarily relied on search and inference techniques developed by the constraint programming community \cite{Beck2010JobShop}. These techniques effectively leverage constraints to define the relationships and limitations between jobs and resources, enabling efficient exploration of feasible solution spaces and the identification of optimal or near-optimal schedules \cite{Nowicki2005TabuSearch}. A widely used heuristic method in real-world scheduling systems is the Priority Dispatching Rule (PDR) \cite{PDR_Zahmani2015JobShop}. PDRs are effective, although designing an efficient PDR is time-consuming and requires extensive domain knowledge.

Recently, approaches utilizing Deep Learning and Neural Networks have gained attention for finding promising solutions to the Job Shop Scheduling Problem (JSSP) \cite{Bonetta2023JobShop, L2D, SelfJSP}. These methods can be broadly categorized into supervised learning and reinforcement learning (RL). Current research in deep reinforcement learning (DRL) is actively focused on developing advanced methods to tackle JSSP. Existing DRL methods typically represent JSSP as a Markov Decision Process (MDP) and learn a policy network based on DRL techniques\cite{L2D}. 

While there are currently no papers that directly address the scheduling of Job Shop Scheduling Problems (JSSP) using LLMs, some notable works explore the potential of LLMs in mathematical reasoning and programming \cite{chen2023program,wei2022chain_of_thaught,ahn2024large,opro}.
Optimization using large language models (LLMs) has gained significant interest in recent years, with several works exploring their capabilities across various domains \cite{opro}. The ability of LLMs to understand and generate natural language has opened new possibilities for optimization tasks that were traditionally solved using derivative-based algorithms or heuristic methods\cite{opro}. Notably, \cite{chen2023program} have done a comprehensive evaluation of LLMs, incorporating an examination of their performance in mathematical problem-solving.  \cite{chen2023program} introduces a novel approach called "Program of Thoughts" (PoT) prompting. Unlike the Chain of Thoughts (CoT) method\cite{wei2022chain_of_thaught}, which uses language models to generate both reasoning steps and computations, PoT separates these tasks. PoT uses language models to generate programming language statements for the reasoning steps and then delegates the actual computation to a program interpreter. 
In \cite{ahn2024large} the authors conduct a comprehensive survey of mathematical problems and corresponding datasets investigated in the context of LLMs. \cite{ahn2024large} examines the spectrum of LLM-oriented techniques for mathematical problem-solving, providing insights into their strengths and weaknesses. \cite{frieder2023large} explores the impact of LLMs on mathematicians' workflows, envisioning changes in research and education through automated assistance and new exploration methods. It provides empirical evidence on LLMs' performance in solving problems and generating proofs, highlighting both successes and failures to give a balanced view of their current capabilities.

More recent works, such as \cite{opro} highlight the potential of LLMs as optimizers, capable of iteratively refining solutions based on a trajectory of previously evaluated solutions. By leveraging the unique strengths of LLMs, such as their natural language understanding and generation capabilities. Paper demonstrates case studies on two fundamental optimization problems: linear regression and the traveling salesman problem. \cite{opro} demonstrates that in small-scale optimization scenarios, LLMs can generate high-quality solutions solely through prompting, sometimes matching or even surpassing the performance of manually crafted heuristic algorithms.

Explorations into using LLMs for graph learning tasks have yielded notable approaches. \cite{chen2024exploring} introduces two pipelines: LLMs-as-Enhancers, where LLMs refine textual data for Graph Neural Networks (GNNs), and LLMs-as-Predictors, where LLMs generate predictions directly from graph structures in natural language. Additionally, \cite{zhao2024graphtext} presents GRAPHTEXT, a method that translates graphs into natural language for LLM-based reasoning. GRAPHTEXT constructs graph-syntax trees for training-free, interactive reasoning, achieving performance on par with or exceeding supervised GNNs through in-context learning, highlighting LLMs' potential in graph machine learning.

\section{Preliminary}

The Job-Shop Scheduling Problem (JSSP) is formally defined as a problem involving a set of jobs \(J\) and a set of machines \(M\). The size of the JSSP problem instance is described as  $N_J \times N_M$, where $N_J$ represents the number of jobs and $N_M$ the number of machines. For each job \(J_i \in J\), it must be processed through \(n_i\) machines (where 
$n_i$ is the number of operations for job $J_i$) in a specified order \(O_{i1} \rightarrow \ldots \rightarrow O_{in_i}\), where each \(O_{ij}\) (for \(1 \leq j \leq n_i\)) represents an operation of \(J_i\) with a processing time \(p_{ij} \in \mathbb{N}\). This sequence also includes a precedence constraint. Each machine can process only one job at a time, and switching jobs mid-operation is not allowed. The objective of solving a JSSP is to determine a schedule, that is, a start time \(S_{ij}\) for each operation \(O_{ij}\), to minimize the makespan \(C_{\max} = \max_{i,j} \{C_{ij} = S_{ij} + p_{ij}\}\) while meeting all constraints. The complexity of a JSSP instance is given by $N_J \times N_M$.

\section{Dataset Generation}
\label{sec:dataset_generation}

In order to try to solve the JSSP with LLM, we first need to represent the problem in natural language. To do that, we have to transform the matrix-based representation in standard JSSP format to a human-readable format. See the example in Listing \ref{lst:problem_instance}.

\begin{lstlisting}[caption={Job Shop Scheduling Problem instance (ft06)\cite{FT06} with $N_J=6$ and $N_M=6$. The problem instance begins with the problem size on the first row, followed by the operations for each job. Odd columns list machines, and even columns list durations. The last row indicates the makespan (55.0)}, label=lst:problem_instance]
6 6        
2 1 0 3 1 6 3 7 5 3 4 6
1 8 2 5 4 10 5 10 0 10 3 4
2 5 3 4 5 8 0 9 1 1 4 7
1 5 0 5 2 5 3 3 4 8 5 9
2 9 1 3 4 5 5 4 0 3 3 1
1 3 3 3 5 9 0 10 4 4 2 1
55.0
\end{lstlisting}

\subsection{Converting JSSP problem instance to Natural Language: Feature Generation}

We use two methods to convert a Job Shop Scheduling Problem (JSSP) from a matrix representation into a human-readable format. Each method presents the information differently. 

\subsection{Approach 1: Job-Centric}
This approach describes the tasks organized by jobs, providing a job-centric view of the scheduling problem.
\begin{itemize}
\item \textbf{Initialization:} Begins by introducing the problem, detailing the number of jobs and machines involved.
\item \textbf{Task Organization:} Enumerates operations for each job, specifying the sequence of operations, the corresponding machines, and their respective durations.
\item \textbf{Description Generation:} Provides a detailed description of each machine's tasks, including the job number, operation number, and duration, ensuring clarity and completeness.
\end{itemize}

\begin{lstlisting}[caption={Job-Centric approach: Natural Language description of a JSSP Problem instance of size $N_J=3$ and $N_M=3$}, label=lst:problem_instance_hl_jcentric]

Optimize schedule for 3 Jobs across 3 Machines to minimize makespan. Each job involves a series of Operations needing specific machines and times. Operations are processed in order, without interruption, on a single Machine at a time.

Problem: 

 Job 0 consists of the following Operations:
  Operation 0 on Machine 0 duration 105 mins.
  Operation 1 on Machine 1 duration 29 mins.
  Operation 2 on Machine 2 duration 213 mins.


 Job 1 consists of the following Operations:
  Operation 0 on Machine 2 duration 193 mins.
  Operation 1 on Machine 1 duration 18 mins.
  Operation 2 on Machine 0 duration 213 mins.


 Job 2 consists of the following Operations:
  Operation 0 on Machine 0 duration 78 mins.
  Operation 1 on Machine 2 duration 74 mins.
  Operation 2 on Machine 1 duration 221 mins.
  
\end{lstlisting}

\subsection{Approach 2: Machine-Centric}
\label{sec:machine-centric} 
\subsubsection*{Overview}
This approach describes the operations organized by machines, providing a machine-centric view of the scheduling problem.

\begin{itemize}
\item \textbf{Initialization:} Begins by introducing the problem, detailing the number of jobs and machines involved.
\item \textbf{Task Organization:} Enumerates operations for each machine, specifying the sequence of operations, the corresponding jobs, and their respective durations.
\item \textbf{Description Generation:} Provides a detailed description of each machine's tasks, including the job number, operation number, and duration, ensuring clarity and completeness.
\end{itemize}


\begin{lstlisting}[caption={Machine-Centric approach: Natural Language description of a JSSP instance of size $N_J=3$ and $N_M=3$}, label=lst:problem_instance_hl_mcentric]

Optimize schedule for 3 Jobs across 3 Machines to minimize makespan. Each job involves a series of Operations needing specific machines and times. Operations are processed in order, without interruption, on a single Machine at a time.

Problem: 

 Machine 0 is used for the following Operations:
  Job 0 Operation 0 duration 105 mins.
  Job 2 Operation 0 duration 78 mins.
  Job 1 Operation 2 duration 213 mins.


 Machine 1 is used for the following Operations:
  Job 0 Operation 1 duration 29 mins.
  Job 1 Operation 1 duration 18 mins.
  Job 2 Operation 2 duration 221 mins.


 Machine 2 is used for the following Operations:
  Job 1 Operation 0 duration 193 mins.
  Job 2 Operation 1 duration 74 mins.
  Job 0 Operation 2 duration 213 mins.
  
\end{lstlisting}

\subsection{Zero-shot inference and Label generation}

Our choice of LLM is relatively small Phi-3-Mini-128K-Instruct open-source model with 128K context size. Later we will refer this model as Phi3. The model is one of the open-source AI models developed by Microsoft. It is a 3.8 billion-parameter, lightweight, state-of-the-art model trained using the Phi-3 datasets. It shows good performance across a variety of language, reasoning, coding, and math benchmarks \cite{abdin2024phi3}. 

Initially, we considered performing zero-shot inference with the Phi3 to solve the JSSP. However, the model consistently produced general descriptions of how to solve the problem instead of actual solutions. Occasionally, it provided partial solutions, but these were mostly infeasible. 


Because the zero-shot inference results were not satisfactory, we decided to finetune the large language model (LLM) using a supervised approach. This required creating a supervised dataset, which included not only the problem formulations in natural language as described in Section \ref{sec:dataset_generation} but also the solutions.

To generate feasible solutions, we employed Google's OR-Tools. The configuration for the Google's OR-Tools solver was set as follows:
\begin{itemize}
    \item Maximum time allowed for the solver: \texttt{300} seconds.
    \item Number of search workers: \texttt{42}.
    \item Search branching strategy: \texttt{cp\_model.AUTOMATIC\_SEARCH}.
\end{itemize}

We have generated approximately 120,000 random JSSP problems of various sizes \footnote{\url{https://github.com/starjob42/datasetjsp}}, ranging from 2x2 to 20x20, with the duration of each operation between 5 and 500 units. We created problems with asymmetric sizes also, such as 3x2 and 10x5, to enhance the model’s generalization capability.
Overall, the final dataset consists of around 120,000 natural language descriptions of JSSP problems along with their feasible solutions. Since we limited the maximum allowed time for Google's OR-Tools to 300 seconds, the optimality of solutions for problems with $N_J >10 $ and $N_M>10$
is not guaranteed.


\begin{lstlisting}[caption={Natural Language description of the solution of JSSP problem instance of size $N_J=3$ and $N_M=3$
}, label=lst:solution_structure]

Solution:

 Job 2 Operation 0 on Machine 0 : 0 + 78 -> 78 
 Job 1 Operation 0 on Machine 2 : 0 + 193 -> 193 
 Job 0 Operation 0 on Machine 0 : 78 + 105 -> 183 
 Job 0 Operation 1 on Machine 1 : 183 + 29 -> 212 
 Job 2 Operation 1 on Machine 2 : 193 + 74 -> 267 
 Job 1 Operation 1 on Machine 1 : 212 + 18 -> 230 
 Job 1 Operation 2 on Machine 0 : 230 + 213 -> 443 
 Job 2 Operation 2 on Machine 1 : 267 + 221 -> 488 
 Job 0 Operation 2 on Machine 2 : 267 + 213 -> 480 

Makespan:  488.0, as it is the maximum end completion time of Operation 2
  
\end{lstlisting}

\section{Training}

To prepare the dataset for fine-tuning we  randomly select an initial user prompt variation from a predefined set to prevent overfitting on a single instruction:
\begin{itemize}
    \item \texttt{"Instruct: Provide a solution schedule for the JSSP problem below, also indicate the makespan."}
    \item \texttt{"Task: Provide the steps of a solution for the JSSP problem and determine the makespan."}
    \item \texttt{"Command: Give a detailed solution to tackle the JSSP problem, focusing on optimizing the makespan."}
\end{itemize}

The context is established by defining the assistant as an expert in JSSP (\{\texttt{"role": "system", "content": "You are an expert in Job Shop Scheduling Problem"}\}). Then the function constructs a message sequence with roles for the system, user, and assistant, incorporating the selected user prompt and corresponding natural language JSSP problem and solution. This sequence is then formatted using a chat template from the tokenizer\cite{huggingface_chat_templating}.

\begin{figure}[!ht]
    \centering
    \includegraphics[width=0.48\textwidth]{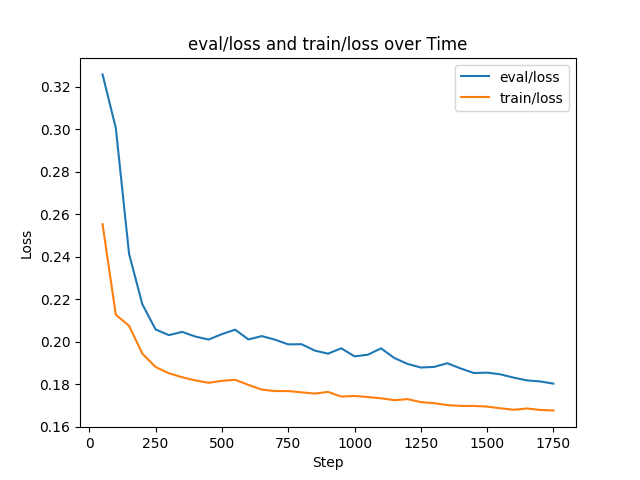}
    \hfill 
    \includegraphics[width=0.48\textwidth]{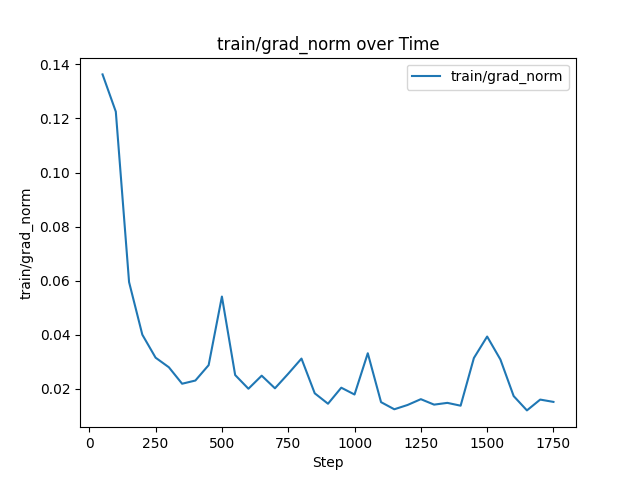}
    \caption{Left: Training and Validation Losses of Phi-3 Model. Right: The Norm of the Gradient during the fine-tuning of Phi-3 Model}
    \label{fig:both_figures}
\end{figure}

The tokenizer is initialized with the following settings: \texttt{padding\_side} is set to 'right', ensuring that padding tokens are added to the end of the sequence; \texttt{pad\_token} is set to the tokenizer's end-of-sequence (\texttt{eos\_token}) token; \texttt{model\_max\_length} is set to 40,000, defining the maximum length of the sequences to avoid memory issues; \texttt{truncation} is enabled to truncate sequences longer than the maximum length; and \texttt{padding} is enabled to pad sequences shorter than the maximum length. These settings ensure that the sequences are appropriately padded and truncated for model input.
In order to fine-tune the model we utilized the LoRA: Low-Rank Adaptation method \cite{hu2022lora}.  The method significantly improves the efficiency of fine-tuning large pre-trained language models. LoRA reduces the number of trainable parameters by injecting trainable rank decomposition matrices into each layer of the Transformer architecture while keeping the pre-trained model weights frozen. 
For fine-tuning, we followed a sample fine-tuning script provided on the Microsoft Hugging Face official website \cite{huggingface_finetuning}.
The base model is loaded with specific quantization configurations to optimize memory and computational efficiency using \textit{BitsAndBytesConfig}, which configures 4-bit quantization with NF4 quantization type. We have used the LoRA rank (128), scaling factor (\textit{lora\_alpha}=256), target modules for adaptation (\textit{qkv\_proj}, \textit{o\_proj}, \textit{fc1}, and \textit{fc2}), and dropout rate (0.05). Gradient checkpointing is enabled to reduce memory usage during training. 
Fine-tuning was performed on a dataset of 120,000 examples as described in Section \ref{sec:dataset_generation}, with 2,000 examples reserved for validation. We used Machine-Centric approach as discussed in \ref{sec:machine-centric} during the training.
We used micro-batch size of 2 and a gradient accumulation step to achieve an effective batch size of 64. The learning rate is set to 2e-4, and the optimizer used is \textit{paged\_adamw\_8bit}. The fine-tuning is set for 1 epoch, with logging and evaluation occurring every 25 steps. We fix the seed to 42 to ensure reproducibility.
The training is performed using single NVIDIA A6000 GPU with 48 GB of RAM for around 242 hours or 10 days.  The training and validation curves are presented on Figure \ref{fig:both_figures}

\section{Evaluation}

To evaluate the model's performance, we created a separate random evaluation dataset distinct from the training data.  This dataset included problems of various sizes, ranging from 2x2 to 9x9, with durations of operations ranging from 1 to 199. around 1000 examples in total. We fix the initial user prompt to \texttt{"Instruct: Provide a solution schedule for the JSSP problem below, also indicate the makespan."}
We observed that performing inference in $float8$ or $float4$ data types resulted in solutions that deviated from the training data format during generation. Therefore, we opted for $float16$ format for testing. Given the memory constraints of a single NVIDIA A6000 GPU with 48GB,
we could only test instances with size $N_J \times N_M < 100$ using $float16$ data type and sampling.
The inference process itself consumes approximately 43GB of memory on the NVIDIA A6000 GPU with $float16$ data type.


\subsection{Overview of JSSP Solution Parsing and Validation}

Following inference, we employ regular expressions to parse the output string generated by the LLM. This process extracts job number, operation number, machine number, start time, duration, end time for each operation, and the makespan value (if present).

\subsubsection{Validating JSSP Solution}
We then use the extracted information from the JSSP solution to perform the following checks:
\begin{itemize}
    \item \textbf{Operation Validation:} Confirms that each operation's machine and duration in the LLM output match the expected values from the problem data.
    \item \textbf{Machine Conflict Check:} Ensures no overlapping operations on the same machine by sorting operations by start time and checking for overlaps.
    \item \textbf{Job Precedence Check:} Verifies that the end time of one operation is before the start time of the next within the same job, ensuring correct operation order.
\end{itemize}
 If all checks pass, the solution is deemed feasible.
\subsection{Hyperparameter Tuning}
We employed specific sampling hyper-parameters during the inference. To determine these hyper-parameters, we conducted a grid search on a small dataset of 7x8 and 8x8 problems, encompassing 200 instances in total.
hyperparameters: 
\begin{itemize}
    \item \texttt{top\_k\_values} = [10, 20, 50]
    \item \texttt{temperature\_values} = [0.2, 0.5, 0.7, 1.0]
    \item \texttt{top\_p\_values} = [0.8, 0.9, 0.95]
\end{itemize}
We fixed \texttt{num\_return\_sequences=10} in all combinations. We found the following configuration to be the best in providing the lowest makespan:

\begin{itemize}
    \item \texttt{sample=True} for sampling-based generation.
    \item \texttt{num\_return\_sequences=10} to generate ten sequences.
    \item \texttt{temperature=1.0} to control randomness.
    \item \texttt{top\_k=50} to limit sampling to the top 50 tokens.
    \item \texttt{top\_p=0.95} to apply nucleus sampling.
\end{itemize}
The tokenizer is configured with a maximum length of 40,000 tokens, uses \texttt{eos\_token} as the padding token, and pads on the left.

\subsection{Comparative Analysis with Other Neural Approaches}

Due to the lack of other works utilizing LLMs end-to-end for scheduling, we compared our results with other neural approaches. We assessed the average gap between the optimal makespan and the makespan of our fine-tuned Phi3 model on our dataset. 

We compared our results with "Learning to Dispatch for Job Shop Scheduling via Deep Reinforcement Learning" (L2D) \cite{L2D}, which uses a Graph Neural Network (GNN) and Proximal Policy Optimization (PPO). L2D's method employs a size-agnostic policy network for generalization. The original paper uses greedy first choice over the policy network's probability distribution. For fair comparison, we sampled the policy network's probability distribution (\textbf{s=10}) and selected the solution with the minimum makespan. We used the network trained on instances with $N_J = 20$ and $N_M = 20$.

Another paper used for comparison is "Self-Labeling the Job Shop Scheduling Problem" (SLJ) \cite{SelfJSP}, which introduces a novel self-supervised training strategy for the JSSP. Their method leverages a generative model based on the Pointer Network, training it by generating multiple solutions (Beta parameter) and using the best one as a pseudo-label, thus eliminating the need for costly optimal solutions. During the comparison, we used three different networks trained with Beta values of 32, 128, and 256, namely slj\_32, slj\_128 and slj\_256 . During testing, the method also utilizes sampling, selecting the best solution from several sampled solutions. We used a sample size of $s=10$ for fair comparison.

\begin{figure}[!ht]
    \centering
    \includegraphics[width=0.8\textwidth]{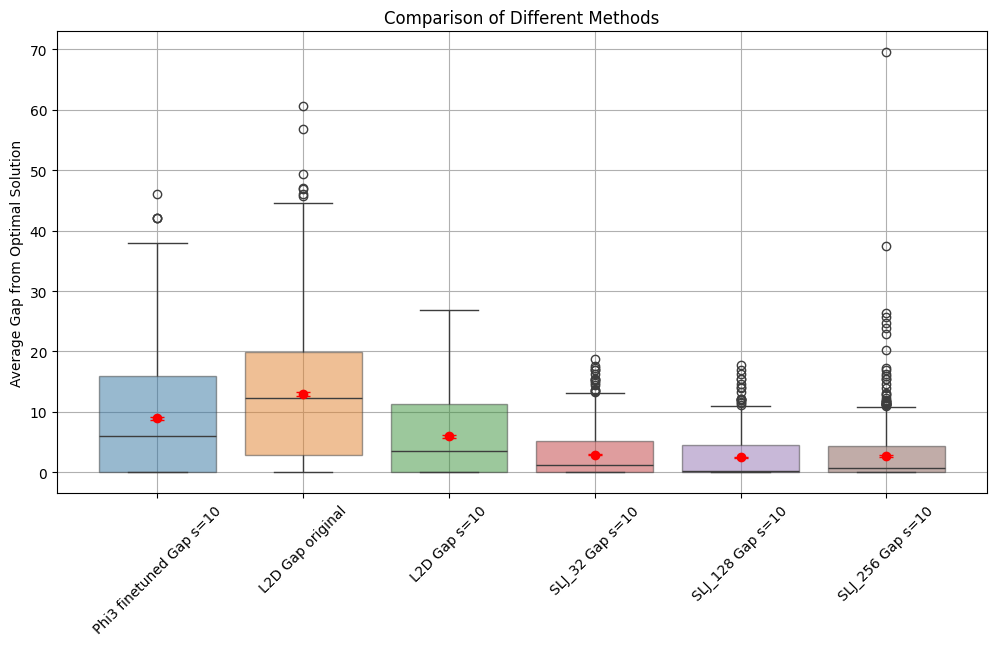}
    \caption{Box plot comparison of different models: (1) L2D original with greedy inference using a 20x20 policy network (L2D Gap original), (2) L2D policy network with sampling (\textbf{s=10}) (L2D Gap s=10), (3) SLJ models trained with $\beta$ values of 32, 128, and 256, and evaluated with sampling (\textbf{s=10}), and (4) Fine-tuned Phi-3 model inference with sampling (\textbf{s=10}). Left side y-axis indicates the gap in percentages and the average time in seconds is indicated by the right y-axis. }
    \label{fig:box_plot}
\end{figure}

Figure \ref{fig:box_plot} shows the comparison of different methods based on the gap from the optimal solution.
From the box plot, it is observed that the Phi3 fine-tuned model shows a relatively low median gap, indicating that it is quite comparable to the other methods. The L2D (original) method has a higher median gap and greater variability, suggesting less consistency compared to the fine-tuned Phi3 and other methods. However, the L2D (sample size $s=10$) method shows a significant improvement over the original L2D, with a lower median gap and reduced variability. The SLJ methods, with Beta values of 32, 128, and 256, demonstrate consistently low median gaps and reduced variability, indicating robust performance across different configurations.

Overall, the fine-tuned Phi-3 model is quite comparable to other methods, given the smaller size of LoRA weights and its capability to perform other language tasks. When fine-tuned on our dataset, the Phi-3 model outperforms the original L2D method. Appropriate sampling significantly enhances the model's performance, reducing the gap to the optimal solution.

\begin{table}
\centering
\caption{This table provides the descriptive statistics for the gap from the optimal solution across different methods: Phi3, L2D, and SLJ with various configurations and with sample size s=10. The metrics include the count, mean, standard deviation, minimum, 25th percentile, median (50th percentile), 75th percentile, and maximum values for each method.}
\label{tab:test_summary_stats} 
\small 
\begin{tabular}{lrrrrrr}
\toprule
{} &  Phi3\_s=10 &  l2d &  l2d\_s=10 &  slj\_32\_s=10 &  slj\_128\_s=10 &  slj\_256\_s=10\\
\midrule
mean  & 8.92 & 13.01 & 5.98 & 2.96 & 2.48 & 2.68 \\
std   & 9.77 & 10.90 & 6.67 & 3.73 & 3.29 & 4.35 \\
min   & 0.00 & 0.00 & 0.00 & 0.00 & 0.00 & 0.00 \\
25\%  & 0.00 & 2.78 & 0.00 & 0.00 & 0.00 & 0.00 \\
50\%  & 6.00 & 12.22 & 3.61 & 1.17 & 0.26 & 0.65 \\
75\%  & 16.00 & 19.88 & 11.25 & 5.30 & 4.48 & 4.34 \\
max   & 46.00 & 60.61 & 26.91 & 18.67 & 17.74 & 69.49 \\
\bottomrule
\end{tabular}
\end{table}

\section{Conclusion}

This paper demonstrates the potential of Large Language Models (LLMs) in addressing the Job Shop Scheduling Problem (JSSP). We introduced a novel supervised dataset with natural language descriptions for JSSP tailored for LLM training. Our results on small-scale JSSP problems indicate that with minimal fine-tuning using the LoRA method\cite{hu2022lora}, Phi-3 can effectively schedule, sometimes matching or surpassing traditional neural network approaches.

The comparative analysis shows that our fine-tuned Phi-3 model performs competitively, achieving an average gap of 8.92 compared to 13.01 for the original L2D method. While the SLJ models (with $\beta$ values of 32, 128, and 256) achieve lower gaps, the Phi-3 model still demonstrates comparable performance, especially considering it is primarily a language model with the ability to handle a wide range of language tasks. Overall, our findings suggest that LLMs, Phi-3 particularly, when enhanced with appropriate sampling methods, offer a promising new avenue for tackling JSSP.

\section{Limitations and Future Work}

Our investigation of the potential of using LLMs for the JSSP shows promising results but also highlights several limitations and future research directions.

A key limitation is the computational overhead of fine-tuning LLMs, which remains resource-intensive. Additionally, due to lack of computational resources the generalizability of our results across larger JSSP instances is uncertain. Further research is needed to test LLMs on larger problem sizes. Comparing the performance of larger LLMs and different fine-tuning methods against the relatively small Phi-3 model is also necessary.

The interpretability of LLM-generated schedules is another challenge, due to their black-box nature. Additionally, while we used a sampling method to improve performance, exploring different sampling strategies could further enhance LLM-generated schedules.

Future research should also explore integrating LLMs with other AI techniques, such as reinforcement learning and graph neural networks, to combine their strengths.

\newpage
\bibliographystyle{plainnat}
\bibliography{main}

\begin{thebibliography}{21}
\providecommand{\natexlab}[1]{#1}
\providecommand{\url}[1]{\texttt{#1}}
\expandafter\ifx\csname urlstyle\endcsname\relax
  \providecommand{\doi}[1]{doi: #1}\else
  \providecommand{\doi}{doi: \begingroup \urlstyle{rm}\Url}\fi

\bibitem[Abdin et~al.(2024)Abdin, Jacobs, Awan, Aneja, Awadallah, Awadalla, Bach, Bahree, Bakhtiari, Bao, Behl, Benhaim, Bilenko, Bjorck, Bubeck, Cai, Cai, Mendes, Chen, Chaudhary, Chen, Chen, Chen, Chen, Chopra, Dai, Giorno, de~Rosa, Dixon, Eldan, Fragoso, Iter, Gao, Gao, Gao, Garg, Goswami, Gunasekar, Haider, Hao, Hewett, Huynh, Javaheripi, Jin, Kauffmann, Karampatziakis, Kim, Khademi, Kurilenko, Lee, Lee, Li, Li, Liang, Liden, Liu, Liu, Liu, Lin, Lin, Luo, Madan, Mazzola, Mitra, Modi, Nguyen, Norick, Patra, Perez-Becker, Portet, Pryzant, Qin, Radmilac, Rosset, Roy, Ruwase, Saarikivi, Saied, Salim, Santacroce, Shah, Shang, Sharma, Shukla, Song, Tanaka, Tupini, Wang, Wang, Wang, Wang, Ward, Wang, Witte, Wu, Wyatt, Xiao, Xu, Xu, Xu, Yadav, Yang, Yang, Yang, Yang, Yu, Yuan, Zhang, Zhang, Zhang, Zhang, Zhang, Zhang, Zhang, and Zhou]{abdin2024phi3}
Marah Abdin, Sam~Ade Jacobs, Ammar~Ahmad Awan, Jyoti Aneja, Ahmed Awadallah, Hany Awadalla, Nguyen Bach, Amit Bahree, Arash Bakhtiari, Jianmin Bao, Harkirat Behl, Alon Benhaim, Misha Bilenko, Johan Bjorck, Sébastien Bubeck, Qin Cai, Martin Cai, Caio César~Teodoro Mendes, Weizhu Chen, Vishrav Chaudhary, Dong Chen, Dongdong Chen, Yen-Chun Chen, Yi-Ling Chen, Parul Chopra, Xiyang Dai, Allie~Del Giorno, Gustavo de~Rosa, Matthew Dixon, Ronen Eldan, Victor Fragoso, Dan Iter, Mei Gao, Min Gao, Jianfeng Gao, Amit Garg, Abhishek Goswami, Suriya Gunasekar, Emman Haider, Junheng Hao, Russell~J. Hewett, Jamie Huynh, Mojan Javaheripi, Xin Jin, Piero Kauffmann, Nikos Karampatziakis, Dongwoo Kim, Mahoud Khademi, Lev Kurilenko, James~R. Lee, Yin~Tat Lee, Yuanzhi Li, Yunsheng Li, Chen Liang, Lars Liden, Ce~Liu, Mengchen Liu, Weishung Liu, Eric Lin, Zeqi Lin, Chong Luo, Piyush Madan, Matt Mazzola, Arindam Mitra, Hardik Modi, Anh Nguyen, Brandon Norick, Barun Patra, Daniel Perez-Becker, Thomas Portet, Reid Pryzant, Heyang
  Qin, Marko Radmilac, Corby Rosset, Sambudha Roy, Olatunji Ruwase, Olli Saarikivi, Amin Saied, Adil Salim, Michael Santacroce, Shital Shah, Ning Shang, Hiteshi Sharma, Swadheen Shukla, Xia Song, Masahiro Tanaka, Andrea Tupini, Xin Wang, Lijuan Wang, Chunyu Wang, Yu~Wang, Rachel Ward, Guanhua Wang, Philipp Witte, Haiping Wu, Michael Wyatt, Bin Xiao, Can Xu, Jiahang Xu, Weijian Xu, Sonali Yadav, Fan Yang, Jianwei Yang, Ziyi Yang, Yifan Yang, Donghan Yu, Lu~Yuan, Chengruidong Zhang, Cyril Zhang, Jianwen Zhang, Li~Lyna Zhang, Yi~Zhang, Yue Zhang, Yunan Zhang, and Xiren Zhou.
\newblock Phi-3 technical report: A highly capable language model locally on your phone, 2024.

\bibitem[Ahn et~al.(2024)Ahn, Verma, Lou, Liu, Zhang, and Yin]{ahn2024large}
Janice Ahn, Rishu Verma, Renze Lou, Di~Liu, Rui Zhang, and Wenpeng Yin.
\newblock Large language models for mathematical reasoning: Progresses and challenges.
\newblock In Neele Falk, Sara Papi, and Mike Zhang, editors, \emph{Proceedings of the 18th Conference of the European Chapter of the Association for Computational Linguistics: Student Research Workshop}, pages 225--237, St. Julian{'}s, Malta, March 2024. Association for Computational Linguistics.
\newblock URL \url{https://aclanthology.org/2024.eacl-srw.17}.

\bibitem[Beck et~al.(2010)Beck, Feng, and Watson]{Beck2010JobShop}
J.~Christopher Beck, T.~K. Feng, and Jean-Paul Watson.
\newblock Combining constraint programming and local search for job-shop scheduling.
\newblock \emph{INFORMS Journal on Computing}, 23\penalty0 (1):\penalty0 1--14, 2010.

\bibitem[Bonetta et~al.(2023)Bonetta, Zago, Cancelliere, and Grosso]{Bonetta2023JobShop}
Giovanni Bonetta, Davide Zago, Rossella Cancelliere, and Andrea Grosso.
\newblock Job shop scheduling via deep reinforcement learning: a sequence to sequence approach.
\newblock \emph{Not Specified}, Aug 2023.

\bibitem[Cebi et~al.(2020)Cebi, Atac, and Sahingoz]{jssp_np}
Ceren Cebi, Enes Atac, and Ozgur~Koray Sahingoz.
\newblock Job shop scheduling problem and solution algorithms: A review.
\newblock In \emph{2020 11th International Conference on Computing, Communication and Networking Technologies (ICCCNT)}, pages 1--7, 2020.
\newblock \doi{10.1109/ICCCNT49239.2020.9225581}.

\bibitem[Chaudhry and Khan(2015)]{performance_of_HeuristicDispatchingRules}
S.~A. Chaudhry and S.~Khan.
\newblock Comparison of dispatching rules in job-shop scheduling problem using simulation: A case study.
\newblock \emph{ResearchGate}, 2015.
\newblock URL \url{https://www.researchgate.net/publication/283505822_Comparison_of_dispatching_rules_in_job-shop_Schedulingproblem_Usingsimulation_A_case_study}.

\bibitem[Chen et~al.(2023)Chen, Ma, Wang, and Cohen]{chen2023program}
Wenhu Chen, Xueguang Ma, Xinyi Wang, and William~W. Cohen.
\newblock Program of thoughts prompting: Disentangling computation from reasoning for numerical reasoning tasks.
\newblock \emph{Transactions on Machine Learning Research}, 2023.
\newblock ISSN 2835-8856.
\newblock URL \url{https://openreview.net/forum?id=YfZ4ZPt8zd}.

\bibitem[Chen et~al.(2024)Chen, Mao, Li, Jin, Wen, Wei, Wang, Yin, Fan, Liu, and Tang]{chen2024exploring}
Zhikai Chen, Haitao Mao, Hang Li, Wei Jin, Hongzhi Wen, Xiaochi Wei, Shuaiqiang Wang, Dawei Yin, Wenqi Fan, Hui Liu, and Jiliang Tang.
\newblock Exploring the potential of large language models (llms) in learning on graphs, 2024.

\bibitem[Corsini et~al.(2024)Corsini, Porrello, Calderara, and Dell'Amico]{SelfJSP}
Andrea Corsini, Angelo Porrello, Simone Calderara, and Mauro Dell'Amico.
\newblock Self-labeling the job shop scheduling problem.
\newblock In \emph{Self-Labeling the Job Shop Scheduling Problem}. Arxiv, 2024.

\bibitem[Face()]{huggingface_chat_templating}
Hugging Face.
\newblock Templates for chat models.
\newblock \url{https://huggingface.co/docs/transformers/en/chat_templating}.

\bibitem[Fisher and Thompson(1963)]{FT06}
Henry Fisher and Gerald~L. Thompson.
\newblock Probabilistic learning combinations of local job-shop scheduling rules.
\newblock In John~F. Muth and Gerald~L. Thompson, editors, \emph{Industrial Scheduling}, chapter 3.2, pages 225--251. Prentice-Hall, Englewood Cliffs, {NJ}, {USA}, 1963.

\bibitem[Frieder et~al.(2024)Frieder, Berner, Petersen, and Lukasiewicz]{frieder2023large}
Simon Frieder, Julius Berner, Philipp Petersen, and Thomas Lukasiewicz.
\newblock Large language models for mathematicians, 2024.

\bibitem[Garey et~al.(1976)Garey, Johnson, and Sethi]{garey1976complexity}
Michael~R Garey, David~S Johnson, and Ravi Sethi.
\newblock The complexity of flowshop and jobshop scheduling.
\newblock \emph{Mathematics of Operations Research}, 1\penalty0 (2):\penalty0 117--129, 1976.

\bibitem[Hu et~al.(2022)Hu, yelong shen, Wallis, Allen-Zhu, Li, Wang, Wang, and Chen]{hu2022lora}
Edward~J Hu, yelong shen, Phillip Wallis, Zeyuan Allen-Zhu, Yuanzhi Li, Shean Wang, Lu~Wang, and Weizhu Chen.
\newblock Lo{RA}: Low-rank adaptation of large language models.
\newblock In \emph{International Conference on Learning Representations}, 2022.
\newblock URL \url{https://openreview.net/forum?id=nZeVKeeFYf9}.

\bibitem[Microsoft()]{huggingface_finetuning}
Microsoft.
\newblock Sample fine-tuning script.
\newblock \url{https://huggingface.co/microsoft/Phi-3-mini-4k-instruct/blob/main/sample_finetune.py}.

\bibitem[Nowicki and Smutnicki(2005)]{Nowicki2005TabuSearch}
Eugeniusz Nowicki and Czeslaw Smutnicki.
\newblock An advanced tabu search algorithm for the job shop problem.
\newblock \emph{Journal of Scheduling}, 8\penalty0 (2):\penalty0 145--159, 2005.
\newblock \doi{10.1007/s10951-005-6364-5}.

\bibitem[Wei et~al.(2022)Wei, Wang, Schuurmans, Bosma, Ichter, Xia, Chi, Le, and Zhou]{wei2022chain_of_thaught}
Jason Wei, Xuezhi Wang, Dale Schuurmans, Maarten Bosma, Brian Ichter, Fei Xia, Ed~H. Chi, Quoc~V. Le, and Denny Zhou.
\newblock Chain-of-thought prompting elicits reasoning in large language models.
\newblock \emph{Google Research, Brain Team}, 2022.

\bibitem[Yang et~al.(2023)Yang, Wang, Lu, Liu, Le, Zhou, and Chen]{opro}
Chengrun Yang, Xuezhi Wang, Yifeng Lu, Hanxiao Liu, Quoc~V Le, Denny Zhou, and Xinyun Chen.
\newblock Large language models as optimizers.
\newblock \emph{arXiv preprint arXiv:2309.03409}, 2023.

\bibitem[Zahmani et~al.(2015)Zahmani, Atmani, Bekrar, and Aissani]{PDR_Zahmani2015JobShop}
Mohamed~Habib Zahmani, Baghdad Atmani, Abdelghani Bekrar, and Nassima Aissani.
\newblock Multiple priority dispatching rules for the job shop scheduling problem.
\newblock In \emph{3rd International Conference on Control, Engineering Information Technology (CEIT’2015)}, Tlemcen, Algeria, 2015.
\newblock \doi{10.1109/CEIT.2015.7232991}.

\bibitem[Zhang et~al.(2020)Zhang, Song, Cao, Zhang, Tan, and Xu]{L2D}
Cong Zhang, Wen Song, Zhiguang Cao, Jie Zhang, Puay~Siew Tan, and Chi Xu.
\newblock Learning to dispatch for job shop scheduling via deep reinforcement learning.
\newblock In \emph{34th Conference on Neural Information Processing Systems (NeurIPS)}, 2020.

\bibitem[Zhao et~al.(2024)Zhao, Zhuo, Shen, Qu, Liu, Bronstein, Zhu, and Tang]{zhao2024graphtext}
Jianan Zhao, Le~Zhuo, Yikang Shen, Meng Qu, Kai Liu, Michael~M. Bronstein, Zhaocheng Zhu, and Jian Tang.
\newblock Graphtext: Graph learning in text space, 2024.
\newblock URL \url{https://openreview.net/forum?id=dbcWzalk6G}.

\end{thebibliography}

\medskip

{
\small

\end{document}